\pdfoutput=1
\documentclass[11pt]{article}

\usepackage[]{ACL2023}

\usepackage{times}
\usepackage{latexsym}

\usepackage[T1]{fontenc}
\usepackage[utf8]{inputenc}

\usepackage{microtype}

\usepackage{inconsolata}

\usepackage{multicol}
\usepackage{booktabs}
\usepackage{makecell}
\usepackage{amsmath, amssymb}
\usepackage{multirow}
\usepackage{mathtools}
\usepackage{xcolor}
\usepackage{enumitem}
\usepackage{subcaption}
\usepackage{float}
\usepackage{graphicx}
\usepackage{caption} 
\usepackage{upgreek}
\usepackage{seqsplit}
\usepackage{color,soul}
\usepackage{arydshln}
\usepackage{placeins}

\definecolor{realblue}{RGB}{0,0,255}

\title{Realistic Conversational Question Answering with Answer Selection \\ based on Calibrated Confidence and Uncertainty Measurement}

\author{Soyeong Jeong$^1$
        \quad Jinheon Baek$^2$
        \quad Sung Ju Hwang$^{1, 2}$
        \quad Jong C. Park$^1$\thanks{\hspace{0.2cm}Corresponding author} \\
        School of Computing$^1$ \quad Graduate School of AI$^2$ \\
        Korea Advanced Institute of Science and Technology$^1$$^,$$^2$\\
       \texttt{\{starsuzi,jinheon.baek,sjhwang82,jongpark\}@kaist.ac.kr}}

\begin{document}
\maketitle

\begin{abstract}
Conversational Question Answering (ConvQA) models aim at answering a question with its relevant paragraph and previous question-answer pairs that occurred during conversation multiple times. To apply such models to a real-world scenario, some existing work uses predicted answers, instead of unavailable ground-truth answers, as the conversation history for inference. However, since these models usually predict wrong answers, using all the predictions without filtering significantly hampers the model performance. To address this problem, we propose to filter out inaccurate answers in the conversation history based on their estimated confidences and uncertainties from the ConvQA model, without making any architectural changes. Moreover, to make the confidence and uncertainty values more reliable, we propose to further calibrate them, thereby smoothing the model predictions. We validate our models, Answer Selection-based realistic Conversation Question Answering, on two standard ConvQA datasets, and the results show that our models significantly outperform relevant baselines. Code is available at: \url{https://github.com/starsuzi/AS-ConvQA}.
\end{abstract}
\section{Introduction}

\begin{figure}[t!]
\begin{center}
\includegraphics[width=0.47\textwidth]{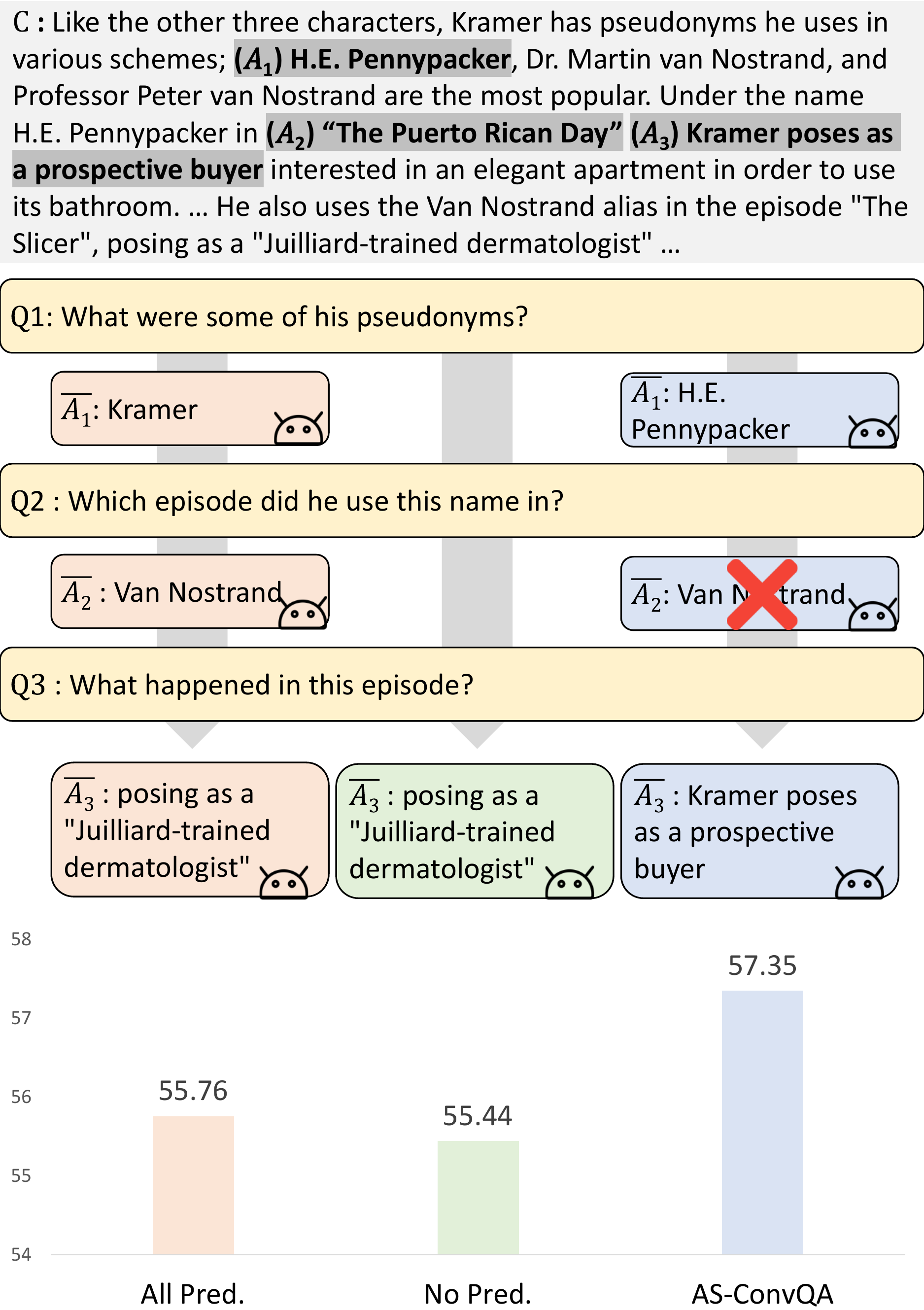}
\end{center}
\vspace{-0.16in}
\caption{
\small Illustration of realistic ConvQA evaluation with three models: 1) using all predicted answers (All Pred.); 2) not using predicted answers (No Pred.); 3) only using probably correct answers while filtering out others (AS-ConvQA, Ours). The scores in the bar chart underneath represent the F1 scores measured by all test samples (see Table~\ref{tab:main} for full results).
% An example and F1-scores of three ConvQA models, All Pred., No Pred, and AS-ConvQA (ours), on a realistic evaluation protocol without using ground-truth answers, $A_1, A_2,$ and $A_3$, as a conversation history to answer a given question, $Q_3$. While All Pred. model uses all predicted answers, $\bar {A_1}$ and $\bar {A_2}$, to answer $Q_3$, No Pred. model does not exploit predicted answers at all. Meanwhile, AS-ConvQA selectively uses the predictions, which helps giving a correct answer, $\bar{A_3}$.
%Note that the ground-truth answers are highlighted in the context, $C$.
}
\label{fig:0_motivation}
\vspace{-0.2in}
\end{figure}

Conversational Question Answering (ConvQA) is the task of answering a series of questions during conversation, taking into account a given relevant paragraph~\cite{convqa1, convqa2}. Contrary to traditional extractive question answering tasks~\cite{SQuAD, NewsQA} that answer each question with the given paragraph just once, ConvQA aims at answering the current question using its previous question-answer pairs taking into account the given paragraph multiple times. For example, as illustrated in Figure~\ref{fig:0_motivation}, the goal of ConvQA is to correctly answer the question $Q_3$ based on the previous conversation history such as $Q_2$, $A_2$, $Q_1$, and $A_1$, as well as the current context $C$.

ConvQA has recently gained much attention as it follows the human's information seeking process through multi-turn interactions with others. However, it is also known to be quite challenging since it requires capturing all the information over the current question, previous conversation, and the given paragraph. To tackle this problem, a considerable amount of work focuses mainly on developing a model architecture for ConvQA~\cite{hae, ham_phae,flowqa,graphflow, excord, DBLP:conf/aaai/QiuHCJQ0HZ21, DBLP:conf/ecir/RaposoRMC22}.

Despite their successes, however, there remains a critical limitation in that they use the ground-truth answers (i.e., $A_2$ and $A_1$) in the conversation history during both training and evaluation steps. Such an evaluation procedure is not applicable to the real-world scenario, since the ground-truth answers are not accessible when the user's query is posed. Therefore, the supporting dialogue history for the current question should consist of the model's predictions $\bar{A_1} $ and $\bar{A_2}$ in the real-world application, instead of the nonexistent gold answers $A_1$ and $A_2$.

There is some recent work~\cite{lrec, acl} that considers such a realistic setting on evaluation. In particular, they propose to use the model's predicted answers (i.e., $\bar{A_1}$ and $\bar{A_2}$), instead of the ground-truth answers (i.e., $A_1$ and $A_2$), for its evaluation.
However, in such a setting, the model faces inconsistency between training and evaluation since the model is evaluated with the predictions while trained with the ground-truth answers. To handle such a discrepancy, \citet{lrec} and \citet{acl} suggest strategies that randomly decide whether to use predicted or gold answers for the input question during training.

However, as Figure~\ref{fig:0_motivation} shows, using all predictions as the answer history is not effective: The performance difference is not so significant when compared to not using them at all. We see that this originates from a model's failure to answering previous questions. Specifically, if a model incorrectly predicts an answer $\bar{A_1}$ for the previous question $Q_1$, using the incorrectly predicted answer $\bar{A_1}$ for the question $Q_2$ not only affects the model's current prediction $\bar{A_2}$ negatively, but also engenders further errors in the future prediction for $Q_3$.

Therefore, in this work, we propose a novel selection scheme for predicted answers from the conversation history, which filters out predictions that are likely to be incorrect, unlike the existing work that uses all the predicted answers including incorrect ones. The remaining step is then to identify possibly incorrect predictions. To this end, we propose to use the confidence and uncertainty of the model's prediction, which are measured by its likelihood and entropy, respectively. In particular, if the model predicts the previous answer with lower confidence (i.e., lower likelihood) or higher uncertainty (i.e., higher entropy) than a certain threshold, we regard the model's previous answer as probably incorrect, and remove it from the conversation history in answering the current question during evaluation. On the other hand, during training, we soften the sampling process so that, instead of using the hard threshold above, we sample a predicted answer based on its confidence or uncertainty (e.g., the lower the uncertainty, the higher the chance to include the predicted answer in the conversation history), in order to diversify the model's input.

However, when dealing with confidence and uncertainty, we should be careful about a miscalibrated situation~\cite{DBLP:conf/icml/GuoPSW17}, which happens when uncertainty and confidence do not correspond to the error and accuracy of ground-truth correctness, respectively. In other words, if the model is not calibrated enough and the distribution for confidence and uncertainty is highly skewed over particular ranges, the highly uncertain or low confident yet valid predictions could be removed. Therefore, to prevent such a performance degrading situation, we further calibrate models using a temperature scaling scheme~\cite{DBLP:conf/icml/GuoPSW17} before estimating the uncertainty or confidence.
We refer to our method as Answer Selection-based realistic Conversational Question Answering (AS-ConvQA). 

We validate our method on two standard ConvQA datasets, QuAC~\cite{convqa1} and CoQA~\cite{convqa2}, against diverse baselines on a realistic evaluation protocol. The experimental results show that our method significantly outperforms these baselines, and a detailed analysis supports the importance of uncertainty- and confidence-based answer selection schemes. 

Our contributions in this work are threefold:
\begin{itemize}[itemsep=0.05mm, parsep=1pt]
  \item We propose to remove incorrect predictions in a conversation history, which degenerate ConvQA models' performances during inference.
  \item We present confidence- and uncertainty-based answer filtering schemes, which are further calibrated to obtain reliable predictions. 
  \item We show that our method achieves outstanding performances on realistic ConvQA tasks.
\end{itemize}
\section{Related Work}

\paragraph{Conversational Question Answering}
ConvQA requires a model to understand the context of questions and paragraphs along with previous conversational questions and answers~\cite{convqa1, convqa2}.
While the simplest approach to consider such conversation histories is to embed them along with the given question and paragraph in the representation space, recent work~\cite{flowqa, ham_phae, graphflow} proposed to leverage the relevant histories by selectively using them. However, \citet{DBLP:conf/wsdm/VakulenkoLTA21} and \citet{excord} have shown that even a simple concatenation of previous questions and answers outperforms these selection-based methods, thanks to the advances in pre-trained language models~\cite{bert, roberta} that are designed to attend to the relevant parts. Furthermore, recent methods~\cite{DBLP:conf/emnlp/ElgoharyPB19, excord, DBLP:conf/wsdm/VakulenkoLTA21, DBLP:conf/ecir/RaposoRMC22} rather focus on the problem of ambiguity in the input question by proposing a question rewriting scheme for its disambiguation, showing remarkable performance improvements.

However, the aforementioned work has a fundamental limitation on the evaluation protocol: They evaluate models based on ground-truth answers working as a conversation history, which are not available in a real-world setting. 
\citet{danqi} point out this problem of ground-truth history evaluation but use the ground-truth answers during evaluation as well, since they target at disambiguating pronouns in the question by comparing the predicted and ground-truth answers.
Alternatively, \citet{lrec} and \citet{acl} use the model's predictions instead of the ground-truth answers during evaluation, and further train the model with predicted answers. 
However, they do not take into account the quality of predicted answers, where low-quality ones are not useful (see Figure~\ref{fig:0_motivation}). Thus, we propose to selectively use the predicted answers that are probably correct, based on their calibrated confidences and uncertainties.

\paragraph{Confidence and Uncertainty}

As it is nearly impossible for models to always make accurate predictions, unreliable predictions become serious issues when deploying machine learning models to real-world settings. Motivated to prevent such a risk, mechanisms of estimating the reliability of model's predictive probabilities based on confidence and uncertainty are recently proposed~\cite{DBLP:journals/inffus/AbdarPHRLGFCKAM21, Houben2022}. We note that confidence is usually measured by the softmax outputs of models~\cite{DBLP:conf/icml/GuoPSW17}, and that uncertainty can be quantified by Bayesian models, which can be approximated via Monte Carlo (MC) dropout~\cite{DBLP:conf/icml/GalG16, DBLP:conf/nips/KendallG17}. With much work on confidence and uncertainty estimations in computer vision tasks~\cite{DBLP:conf/iccv/GuillorySEDS21}, related topics have been recently adopted for NLP tasks as well~\cite{DBLP:conf/eacl/ShelmanovPKBLKK21, DBLP:conf/emnlp/WuLZLHL21, DBLP:conf/iclr/MalininG21, DBLP:conf/acl/VazhentsevKSTTF22}.
While confidence and uncertainty estimation should also be considered in ConvQA, we believe that this venue is under-explored so far. In particular, since questions are asked sequentially, it is likely that untrustworthy predictions in the conversation history would negatively affect the performance. To tackle this, we propose to exclude low-confident or uncertain predictions when training and evaluating the ConvQA model.

\paragraph{Calibration}

Confidence and uncertainty help interpret the validity of the model's prediction. However, it is not safe to rely on them when the model is not calibrated, where the correct likelihood does not match the predicted probability~\cite{DBLP:conf/icml/GuoPSW17}, or the model error does not match the predicted uncertainty~\cite{Laves2019WellcalibratedMU}. Since deep neural networks are prone to miscalibration as the number of parameters has much increased, large pre-trained language models are also not free from this problem~\cite{DBLP:conf/nips/WangLSY21, DBLP:conf/icml/ZhaoWFK021, DBLP:conf/emnlp/DanR21}. One of the most prevalent approaches to calibrating the model is to rescale a logit vector before the softmax function for regularizing the probability, which is known as temperature scaling~\cite{DBLP:conf/icml/GuoPSW17}. While there exist lots of calibration schemes, including label smoothing~\cite{DBLP:conf/cvpr/SzegedyVISW16} and confidence penalty~\cite{DBLP:conf/iclr/PereyraTCKH17}, in this work, we use temperature scaling as a calibrator, since it is simple yet effective while not changing the output class of the model prediction (i.e., only scaling logits).
\section{Method}
\label{sec:method}

We first introduce ConvQA. Then, we describe our answer validating methods based on confidence and uncertainty values with their calibration schemes.

\subsection{Conversational Question Answering}
\label{subsec:convqa}

We provide general descriptions of a ConvQA task. For the $i$-th turn of the conversation, we are given a question $Q_i$ and its corresponding context $C$, as well as its conversation history consisting of previous questions and answers: $\mathcal{H}_i = \{ Q_{i-1}, A_{i-1}, ... , Q_1, A_1 \}$. Then, the goal of ConvQA is to correctly extract the ground-truth answer $A_{i}$ from $C$ along with $Q_i$ and $\mathcal{H}_i$, as follows: 
\begin{equation}
\fontsize{9.5pt}{9.5pt}\selectfont
   P(A_i) = M_{\theta} (C, Q_i, Q_{i-1}, A_{i-1}, ... , Q_1, A_1 ),
\label{eq:existing_convqa}
\fontsize{9.5pt}{9.5pt}\selectfont
\end{equation}
where $M_{\theta}$ is a ConvQA model, parameterized by $\theta$, and, for the sake of simplicity, we omit conditional variables $C$, $Q_i$, and $\mathcal{H}_i$ on the left side of Equation~\ref{eq:existing_convqa}, i.e., $P(A_i) = P(A_i | C, Q_i, \mathcal{H}_i)$. Note that existing work~\cite{DBLP:conf/emnlp/ElgoharyPB19, excord, DBLP:conf/wsdm/VakulenkoLTA21, DBLP:conf/ecir/RaposoRMC22} has an unrealistic assumption that a set of ground-truth answers $\{ A_{i-1}, ..., A_1 \}$ is available during evaluation as in Equation~\ref{eq:existing_convqa}. 
However, this evaluation setup is far from reality, since they are not always available when the user's novel questions come in, unlike the training phase which optimizes a model with ground-truth answers. 
Therefore, we should particularly modify the formulation in Equation~\ref{eq:existing_convqa} to accommodate a realistic evaluation scenario, which we describe in the next subsection. 

\subsection{Realistic ConvQA}
\label{subsec:realistic}

To tackle the problem of accessing ground-truth answers during evaluation in Equation~\ref{eq:existing_convqa}, we aim at redefining its formulation to evaluate ConvQA models under real-world situations as shown below.

\paragraph{Evaluation}

When a user asks a unique question whose ground-truth answers are not accessible, the most naïve approach is to work with the relevant context and previous questions, as follows: 
\begin{equation}
\fontsize{9.5pt}{9.5pt}\selectfont
\hspace*{-3mm}
  P(\bar{A_i}) = M_{\theta} (C, Q_i, Q_{i-1}, ... , Q_1),
\label{eq:noanswer}
\hspace{-3mm}
\fontsize{9.5pt}{9.5pt}\selectfont
\end{equation}
where $\bar{A_i}$ denotes the $i^{th}$ predicted answer $( i > 1 )$ during inference time. However, the formulation in Equation~\ref{eq:noanswer} may be suboptimal, since it ignores predicted answers $\{ \bar{A_{i-1}}, ..., \bar{A_1} \}$ that occurred in the former conversation, which may be beneficial for the current prediction. Thus, we can instead make inference with predicted answers, as follows:
\begin{equation}
\fontsize{9.5pt}{9.5pt}\selectfont
\hspace*{-3mm}
  P(\bar{A_i}) = M_{\theta} (C, Q_i, Q_{i-1}, \bar{A}_{i-1}, ... , Q_1, \bar{A_1} ).
\label{eq:realistic_convqa_eval}
\hspace{-3mm}
\fontsize{9.5pt}{9.5pt}\selectfont
\end{equation}
However, when evaluating with Equation~\ref{eq:realistic_convqa_eval} while training with Equation~\ref{eq:existing_convqa}, a problematic discrepancy arises, as model $M_{\theta}$ uses gold answers $A_i$ for training but predicted answers $\bar{A_i}$ for inference.

\paragraph{Training}

To tackle this inconsistency, recent work~\cite{lrec, acl} randomly decides whether to use $\bar{A_{i}}$ or $A_{i}$ during the training phase, as follows:
\begin{equation}
\fontsize{9pt}{9pt}\selectfont
    P(A_i) =
        \begin{cases}
        M_{\theta} (C, Q_i, Q_{i-1}, \bar{A}_{i-1}, ... ) \; \text{w.p. $\lambda_{rand}$}, \\
        M_{\theta} (C, Q_i, Q_{i-1}, A_{i-1}, ... ) \; \text{w.p. $1 - \lambda_{rand}$},
        \end{cases}
\label{eq:realistic_convqa_train_acl_lrec}
\fontsize{9pt}{9pt}\selectfont
\end{equation}
where $\lambda_{rand}$ is the probability of using $\bar{A_i}$, which is set based on heuristic sampling schemes, either using the random coin flipping or increasing the sampling rate based on the number of steps.

While such an attempt bridges the gap between training and inference in the real-world setting, critical limitations remain.
First, as Figure~\ref{fig:0_motivation} shows, we observe that using all the predicted answers rarely contributes to the model performance, as they include incorrect answers that hinder accurate predictions for the current question.
Also, we further point out that there still exists a discrepancy between training and evaluation: The model observes ground-truth answers in Equation~\ref{eq:realistic_convqa_train_acl_lrec} which are yet unobservable for evaluation in Equation~\ref{eq:realistic_convqa_eval}. Therefore, to tackle these challenges, we propose to selectively use the predicted answers based on the predictions' confidences and uncertainties.
\subsection{Predicted Answer Selection Scheme}
\label{subsec:selection}

Our key intuition is that confidence and uncertainty are simple yet effective measures to filter out inaccurate predictions. Before going into details, we first define the notations. Let $\boldsymbol{x_{i}} \in X$ be an $i^{th}$ input (i.e., turn) for ConvQA model $M_\theta$, which consists of current question $Q_i$, its relevant context $C$, and conversation history $\mathcal{H}_i$. Then, labels of given input $\boldsymbol{x_{i}}$ are defined as $y^{(start)}_i \in C$ and $y^{(end)}_i \in C$ with $C \in \{ 1, ..., K \}$, where $K$ is the number of sequence lengths for context $C$. In other words, $y^{(start)}_i$ and $y^{(end)}_i$ denote the start and end spans, respectively. Further, to predict labels $y^{(start)}_i$ and $y^{(end)}_i$, we first obtain a logit vector $\boldsymbol{z}_{i}$ for each label\footnote{We omit superscripts $start$ and $end$ for simplicity.}, and use it for calculating a probability vector $\boldsymbol{p}_i$ over $K$ spans: $\boldsymbol{p}_i = \texttt{softmax} (\boldsymbol{z}_{i})$, where $\texttt{softmax}$ is a softmax function.

\paragraph{Confidence}
We now define the confidence. From the probability $\boldsymbol{p} = \texttt{softmax} (\boldsymbol{z})$, a model likelihood can be interpreted as confidence, as follows\footnote{For simplicity, we omit a turn index $i$, which is represented in a subscript, for example, $Q_i$ for the $i^{th}$ conversation.}:
\begin{equation}
    s_{conf} = \max_{y \in C} p(y | \boldsymbol{z}),
\label{eq:confidence}
\end{equation}
where $s_{conf}$ denotes the confidence value.

\paragraph{Uncertainty}
While confidence can estimate how confident the model is on its prediction, it might be also beneficial to measure the model's certainty with Bayesian deep learning techniques~\cite{DBLP:conf/nips/KendallG17} to prevent erroneous predictions, which we describe here. At first, to calculate the uncertainty value, we need to obtain $N$ different predictions for approximating the model's distribution. To do so, we first enable dropout~\cite{DBLP:journals/jmlr/SrivastavaHKSS14} in the language model during inference, and then forward input $\boldsymbol{x}$ for $N$ times with $N$ different dropout masks, which is referred to as Monte Carlo (MC) dropout~\cite{DBLP:conf/icml/GalG16}. Then, we can obtain probability vector $\boldsymbol{p}$ via MC integration: $\boldsymbol{p} = \frac{1}{N} \sum_{n = 1}^N \texttt{softmax} (\boldsymbol{z}^{(n)})$, where $\boldsymbol{z}^{(n)}$ is the logit vector from each forward pass. Then, based on probability $\boldsymbol{p}$, the uncertainty is quantified via its entropy over $K$ classes~\cite{DBLP:conf/nips/KendallG17, Laves2019WellcalibratedMU}, as follows:
\begin{equation}
   s_{uncer} = - \frac{1}{\log K} \sum_{k = 1}^K {p(k)} \log {p(k)},
\label{eq:uncertainty_entropy}
\end{equation}
where $s_{uncer}$ denotes the uncertainty value, which we normalize to be on a scale between 0 and 1 with $\frac{1}{\log K}$ in Equation~\ref{eq:uncertainty_entropy}, following~\cite{Laves2019WellcalibratedMU}.

\subsection{Calibrating Confidence and Uncertainty}
\label{subsec:calibration}
We then describe the calibration schemes to match the model's predicted confidence and uncertainty to its correct likelihood and error, respectively.

\paragraph{Perfect Calibration}
In order to calibrate trustworthiness of the confidence and uncertainty, we first describe perfectly calibrated situations. Given the input $\boldsymbol{x}$, the model predicts the most likely class, $\bar{y} = \arg\max \boldsymbol p$, from the entire classes with the highest probability, $\bar{p} = \max \boldsymbol p$. Each perfect calibration for confidence and uncertainty is then as follows~\cite{DBLP:conf/icml/GuoPSW17, Laves2019WellcalibratedMU}:
\begin{equation}
\begin{split}
   & \mathop{\mathbb{P}} (\bar{y} = y | s_{conf} = p) = p, \\
   & \mathop{\mathbb{P}} (\bar{y} \neq y | s_{uncer} = p) = p,
\end{split}
\label{eq:perfect_calibration}
\end{equation}
where $y$ denotes the true label with $\forall p \in [0,1]$.

\paragraph{Calibration and Uncertainty Error}
However, perfect calibration defined in Equation~\ref{eq:perfect_calibration} is hardly achievable in practical settings due to noise and prediction errors. Thus, we rather define a calibration error to estimate how much the model's prediction is calibrated. One of the most prevalent methods to quantify calibration error for confidence is to measure the difference in expectation between confidence and accuracy as follows~\cite{DBLP:conf/icml/GuoPSW17}:
\begin{equation}
    \mathop{\mathbb{E}_{s_{conf}}}[ \; | \mathop{\mathbb{P}} (\bar{y} = y | s_{conf} = p) - p | \; ],
\label{eq:expectation_calibration_confidence}
\end{equation}
where $\forall p \in [0,1]$. Also, miscalibration of uncertainty is quantified as follows~\cite{Laves2019WellcalibratedMU}:
\begin{equation}
    \mathop{\mathbb{E}_{s_{uncer}}}[ \; | \mathop{\mathbb{P}} (\bar{y} \neq y | s_{uncer} = p) - p | \; ].
\label{eq:expectation_calibration_uncertainty}
\end{equation}

However, since $s_{conf}$ and $s_{uncer}$ lie in a continuous domain, it is impossible to sample them infinite times for every $p$ when measuring calibration errors. Therefore, we further approximate them in a discrete space, which was in the continuous domain (Equations~\ref{eq:expectation_calibration_confidence},~\ref{eq:expectation_calibration_uncertainty}), by dividing the predictions into $M$ bins and then measuring accuracy for each corresponding bin. Formally, accuracy and confidence per bin are as follows~\cite{DBLP:conf/icml/GuoPSW17}:
\begin{equation}
\fontsize{10pt}{10pt}\selectfont
\begin{split}
        & \texttt{acc}(B_m) = \frac{1}{|B_m|} \sum_{i \in B_m} \boldsymbol{1} (s_{conf} = y^{(i)}),\\
        & \texttt{conf}(B_m) = \frac{1}{|B_m|} \sum_{i \in B_m} s_{conf},
\end{split}
\fontsize{10pt}{10pt}\selectfont
\label{eq:bin_confidence}
\end{equation}
where $B_m$ is a set of label indices whose values are within the $m^{th}$ bin among $M$ non-overlapping bins. Similarly, error and uncertainty per bin are formally defined as follows:
\begin{equation}
\fontsize{10pt}{10pt}\selectfont
\begin{split}
        & \texttt{err}(B_m) = \frac{1}{|B_m|} \sum_{i \in B_m} \boldsymbol{1} (s_{uncer} \neq y^{(i)}),\\
        & \texttt{uncer}(B_m) = \frac{1}{|B_m|} \sum_{i \in B_m} s_{uncer}.
\end{split}
\fontsize{10pt}{10pt}\selectfont
\label{eq:bin_uncertainty}
\end{equation}

Using definitions in Equations~\ref{eq:bin_confidence},~\ref{eq:bin_uncertainty} above, we now measure the approximated calibration errors. Regarding confidence, the Expected Calibration Error (ECE)~\cite{DBLP:conf/icml/GuoPSW17} is defined as follows:
\begin{equation}
\fontsize{10pt}{10pt}\selectfont
        ECE = \sum_{m=1}^{M} \frac{|B_m|}{n} |\texttt{acc}(B_m) - \texttt{conf}(B_m)|,
\fontsize{10pt}{10pt}\selectfont
\label{eq:ece}
\end{equation}
where $n$ is the number of samples in total. For uncertainty, Expected Uncertainty Calibration Error (UCE)~\cite{Laves2019WellcalibratedMU} is defined as follows:
\begin{equation}
\fontsize{10pt}{10pt}\selectfont
        UCE = \sum_{m=1}^{M} \frac{|B_m|}{n} |\texttt{err}(B_m) - \texttt{uncer}(B_m)|.
\fontsize{10pt}{10pt}\selectfont
\label{eq:uce}
\end{equation}

\paragraph{Calibration with Temperature Scaling}
With the calibration criteria (i.e., $ECE$ and $UCE$), we now aim at obtaining well-calibrated confidence and uncertainty values having low ECE and UCE. To do so, we apply a temperature scaling scheme, which regulates the scale of the obtained logit vector $\boldsymbol{z}$ with a single scalar, namely temperature $\tau > 0$. Note that temperature scaling does not affect the maximum value of the softmax output; therefore, accuracy is preserved. Formally, the calibrated probability vector $\boldsymbol{\hat{p}}$ is defined as follows:
\begin{equation}
    \boldsymbol{\hat{p}} = \texttt{softmax} (\boldsymbol{z} / \tau).
\label{eq:confidence_temperature_scaling}
\end{equation}
We find the $\tau$ value based on the low calibration errors, i.e., $ECE$ and $UCE$, in experiments.

\subsection{Overall Pipeline}
\label{subsec:overall}

We now summarize the overall pipeline of our AS-ConvQA framework, which leverages the calibrated confidence and uncertainty values to sample valid predictions for inference, while using them during training as well. Our training pipeline consists of two steps, which we explain below.
\paragraph{Step 1}
We start training a model with gold answers $A_i$ following the training protocol in Equation~\ref{eq:existing_convqa}, since, if the model cannot observe gold answers, it might fail to capture and generate accurate answers, easily leading to degenerated performances~\cite{lrec}.
Then, to prepare for Step 2, we make inference with it to obtain prediction $\bar A_i$ together with its confidence and uncertainty, for each input $\boldsymbol{x}_i$ in the training set.

\paragraph{Step 2}
With the predicted answers and their confidences and uncertainties from Step 1, we further train the model to reflect the predicted answers instead of the ground-truth answers. Note that our objective is to filter out less confident or uncertain predictions in inference. Thus, since filtered ones are not observable during our realistic evaluation phase, we also aim at reflecting such an occurrence during training to narrow the gap between training and evaluation. To do so, instead of training with all predicted answers, we rather sample a predicted answer based on its confidence or uncertainty value:
\begin{equation}
\fontsize{9pt}{9pt}\selectfont
    P(A_i) =
        \begin{cases}
        M_{\theta} (C, Q_i, Q_{i-1}, \bar{A}_{i-1}, ... ) \; \text{w.p. $\lambda_{valid}$}, \\
        M_{\theta} (C, Q_i, Q_{i-1}, ... ) \; \text{w.p. $1 - \lambda_{valid}$},
        \end{cases}
\label{eq:realistic_convqa_train_ours}
\fontsize{9pt}{9pt}\selectfont
\end{equation}
where $\lambda_{valid}$ is obtained by the previous prediction's ($\bar{A}_{i-1}$) confidence or uncertainty: $\lambda_{valid} \in [s_{conf}, 1 - s_{uncer}]$. Note that, in contrast to existing work~\cite{lrec, acl} represented in Equation~\ref{eq:realistic_convqa_train_acl_lrec}, our work does not use previous gold-answers ($A_{i-1}$) for training as well.

For evaluation, we follow the realistic evaluation protocol described in Equation~\ref{eq:realistic_convqa_eval}. However, instead of using all predictions~\cite{lrec, acl}, we rather remove low-confident or uncertain predictions against the threshold.

\section{Experimental Setups}

We explain datasets, metric, and models. Please see Appendix~\ref{appendix:implementation_details} for further implementation details.

\subsection{Dataset and Metric}

\aboverulesep=0ex
\belowrulesep=0ex

\begin{table}
\centering
\begin{center}

\resizebox{0.49\textwidth}{!}{
\begin{tabular}[h]{l|cccc}
%\Xhline{2\arrayrulewidth}
\toprule
\multicolumn{1}{c|}{} & \multicolumn{2}{c}{\textbf{QuAC}} & \multicolumn{2}{c}{\textbf{CoQA}} \\
\cmidrule(l{2pt}r{2pt}){2-3} \cmidrule(l{2pt}r{2pt}){4-5} 
%\noalign{\smallskip}\noalign{\smallskip}\hline\hline
\multicolumn{1}{c|}{} & \textbf{BERT} & \textbf{RoBERTa} & \textbf{BERT} & \textbf{RoBERTa} \\
%\hline
\midrule
\multirow{1}{*}{Gold}
 & 59.86 &  65.08 & 72.79 & 77.62 \\
\hline
\multirow{1}{*}{No Pred.}
 & 55.44  & 61.24 & 70.83 & 75.56 \\
\multirow{1}{*}{All Pred.}
 & 55.76 &  61.53 & 71.28 & 75.42 \\
 \multirow{1}{*}{CoQAM}
 & 55.83 & 61.55 & 71.27 & 74.29 \\
 \multirow{1}{*}{Robust-P}
 & 54.21  &  60.32 & 70.17 & 73.96 \\
 \multirow{1}{*}{Attentive Selection}
 & 55.74  &  61.42 & 71.05 & 74.60 \\
 \hline
 \multirow{1}{*}{AS-ConvQA$_\texttt{conf}$ (Ours)}
 & \textbf{57.03} &  \textbf{62.47} & \textbf{72.00} & \textbf{76.52} \\
 \multirow{1}{*}{AS-ConvQA$_\texttt{uncer}$ (Ours)}
 & \textbf{57.35}  & \textbf{62.33} & \textbf{72.08} & \textbf{76.33} \\
 \hline
 \multirow{1}{*}{AS-ConvQA$_\texttt{combine}$ (Ours)}
 & \textbf{57.06} &  \textbf{62.18} & \textbf{71.99} & \textbf{76.76} \\
\bottomrule
\end{tabular}
}
\end{center}
\vspace{-0.15in}
\caption{\small F1-scores on QuAC and CoQA. Note that Gold model is not a fair baseline as it uses the ground-truth answers during inference, and thus is evaluated in an unrealistic setting. 
%The best top-3 models are emphasized in \textbf{bold}.
}
\label{tab:main}
\vspace{-0.15in}
\end{table}

\paragraph{QuAC}

QuAC~\cite{convqa1} is the benchmark ConvQA dataset, which is known to resemble a realistic information seeking dialogue, where questioners were prevented from reading paragraphs for its collection. QuAC consists of 14K dialogues and 100K pairs of questions and paragraphs. As the test set is not publicly open, we use a development set.

\paragraph{CoQA}
CoQA~\cite{convqa2} is another ConvQA dataset with 127K pairs of questions and paragraphs; however, unlike QuAC, questioners were allowed to share paragraphs during collection. We also use a development set instead of the test set, which is not publicly available.

\paragraph{F1-score}
We evaluate models with F1-score, following the standard protocol~\cite{excord}.

\subsection{Question Answering Models}
For question answering models, we use two base-size pre-trained language models widely used in ConvQA tasks: \textbf{BERT-base}~\cite{bert} and \textbf{RoBERTa-base}~\cite{roberta}.

\subsection{Baselines and Our Models}
We compare AS-ConvQA to other relevant baselines using predicted answers. Gold model, which is an indicator, uses gold answers as the answer history during evaluation, which is not realistic, whereas all the others are evaluated with predicted answers. All models are trained with the same protocol, using gold answers as the conversation history for the first half of training epochs (Step 1).

\paragraph{Gold}
This model uses the ground-truth answers during training and evaluation, thus unrealistic.

\paragraph{No Prediction (No Pred.)}
This model does not use the predicted answers as the conversation history in either training or evaluation steps. 

\paragraph{All Prediction (All Pred.)}
In contrast to No Pred., this model uses all the predicted answers during both training and evaluation steps.

\paragraph{CoQAM}
For training, this model uses the random sampling scheme represented in Equation~\ref{eq:realistic_convqa_train_acl_lrec}, which samples either predicted or ground-truth answers with coin-flipping~\cite{lrec}. For evaluation, it uses all the predictions as the history.

\paragraph{Robust-P}
Similar to CoQAM, this model uses a heuristic answer sampling scheme in a random manner, but increases the predicted answer sampling rate for training~\cite{acl}. Also, it is evaluated with all predicted answers.

\paragraph{Attentive Selection}
This model uses the attention mechanism to softly select the relevant answers in the history, following previous work~\cite{ham_phae, flowqa, graphflow}.

\paragraph{AS-ConvQA$_\texttt{conf}$ (Ours)}
This is our model that filters out unconfident answers via confidence values during training and evaluation, after calibration.

\paragraph{AS-ConvQA$_\texttt{uncer}$ (Ours)}
This is also our model that filters out uncertain answers during training and evaluation, after calibrating uncertainty values.

\paragraph{AS-ConvQA$_\texttt{combine}$ (Ours)}
This model combines our confidence and uncertainty modules, where we use the mean of calibrated confidence and $($1-uncertainty$)$ values for filtering out samples.

\begin{figure}[t!]
\begin{center}
\includegraphics[width=0.48\textwidth]{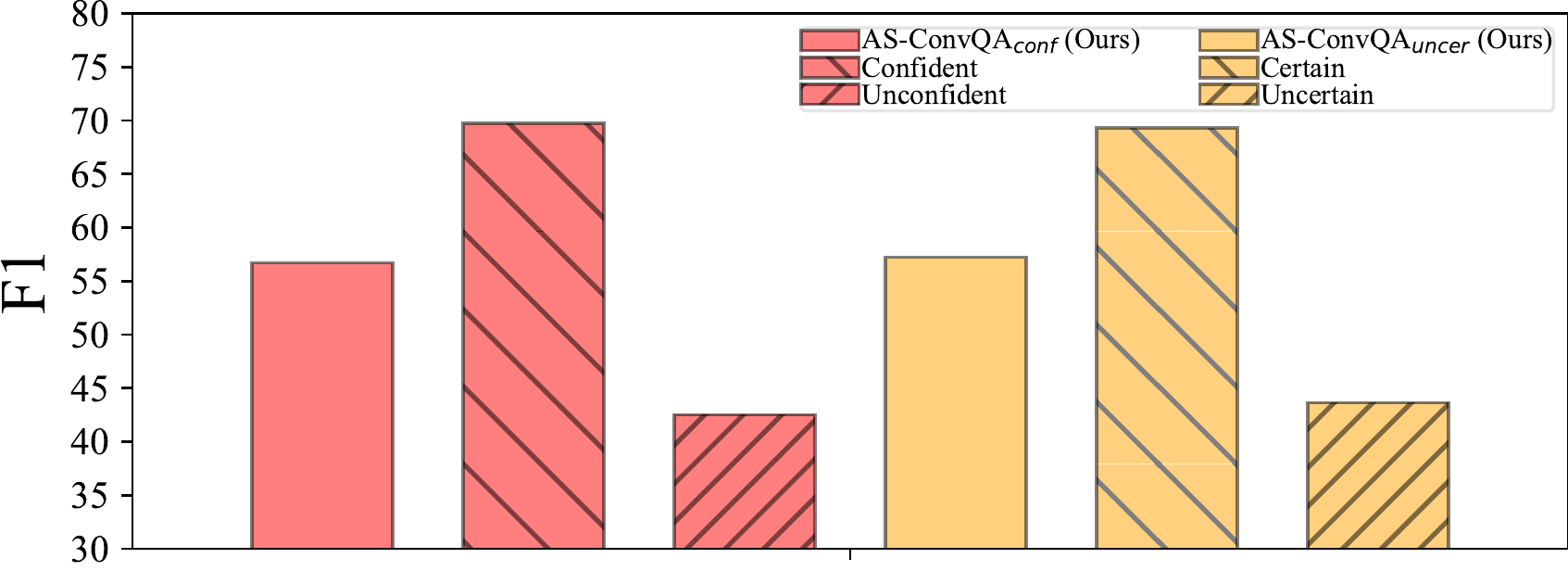}
\end{center}
\vspace{-0.18in}
\caption{
\small Comparison results of certain (confident) and uncertain (unconfident) predictions on QuAC. Note that a threshold is set as the median of the uncertainty (confidence) values.
}
\label{fig:conf_uncer_bar}
\vspace{-0.18in}
\end{figure}

\section{Results and Discussion}
\label{sec:results}
In this section, we show overall performances of our proposed method along with detailed analyses.

\paragraph{Main Results}

As Table~\ref{tab:main} shows, the proposed AS-ConvQA models including confidence and uncertainty schemes show significant performance gains over all baselines on two different QA models. Interestingly, No Pred. model, which does not utilize previous answers as the conversation history, shows comparable to or even better performance than the other baseline models based on either exploiting all the predicted answers or randomly sampling them with heuristic ratios. This implies that it is more helpful not to use low-quality predicted answers -- unconfident or uncertain -- at all than to use them. On the other hand, our models take advantage of filtering out probably invalid predictions, thus achieving improved performance.

Moreover, our AS-ConvQA models outperform the attention-based history selection model (i.e., Attentive Selection). This is because, even though previous answers are all incorrect, the attention scheme should leverage some of them (i.e., the sum of attention scores for previous answers should be 1), which leads the model to answer with an inaccurate history. Meanwhile, AS-ConvQA models can ignore possibly wrong predictions, thus decreasing the risk of being affected by the inaccurate history.

Last, when combining confidence and uncertainty modules, the performance is not much further enhanced. To analyze this, we first measure the number of overlapping questions, where each of the AS-ConvQA$_\texttt{conf}$ and AS-ConvQA$_\texttt{uncer}$ models predicts with higher confidence or lower uncertainty than its median value. Then, we observe that about 74.82\% and 77.12\% of the questions overlap on QuAC and CoQA, respectively. This indicates that unconfident and uncertain samples are highly correlated, which are likely to be filtered out by both confidence- and uncertainty-based models. In other words, due to similar effects of AS-ConvQA$_\texttt{conf}$ and AS-ConvQA$_\texttt{uncer}$ models, the performance of combined models is not much improved.

\begin{figure}[t!]
\begin{center}
\vspace{0.02in}
\includegraphics[width=0.48\textwidth]{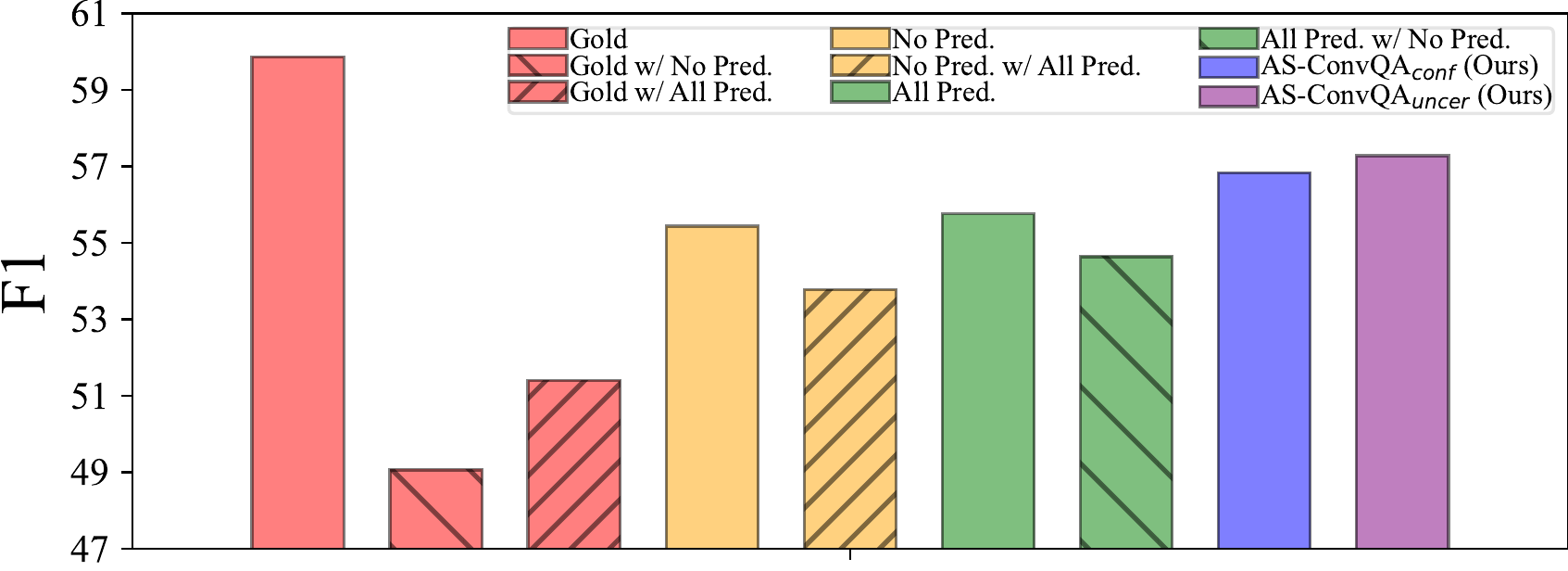}
\end{center}
\vspace{-0.18in}
\caption{
\small F1-scores on the mismatching evaluation settings for each baseline model on QuAC, either using all of the predictions or none of them as the previous answer history.
}
\label{fig:mismatch_bar}
\vspace{-0.18in}
\end{figure}

\paragraph{Unconfident and Uncertain Predictions}
In order to see whether predictions with low confidence or high uncertainty actually correspond to incorrect answers, we compare the performances between the certain (unconfident) and uncertain (confident) predictions. As Figure~\ref{fig:conf_uncer_bar} shows, low-confident and uncertain samples lead to drastic performance degradation. This result corroborates our hypothesis that a prediction with low confidence or high uncertainty acts as an obstacle in ConvQA tasks.

\paragraph{Impact of Realistic Evaluation Setups}
To see results in realistic settings -- not using ground-truth answers during inference -- for the Gold model, we train it with ground-truth answers, and then test either with predicted answers or without them. Figure~\ref{fig:mismatch_bar} shows that performances of the Gold model are drastically dropped, and even lower than both No Pred. and All Pred., even though tested on the same strategies. This can be explained with the term of exposure bias~\cite{DBLP:conf/nips/BengioVJS15, lrec}, where a discrepancy exists between training and evaluation, which hinders the model from performing well on test data that differs from training data. Furthermore, this also explains one of the reasons why CoQAM and Robust-P models perform poorly: Since they observe ground-truth answers for training, which are not observable during evaluation, they underperform ours.

\paragraph{Training \& Evaluation Discrepancy}
We have observed a discrepancy between training and evaluation for the Gold model above. Then, the next possible question is whether this discrepancy also happens for models that are trained on the predicted answers, but evaluated in different settings. To see this, we test No Pred. and All Pred. models in a mismatching evaluation setting. As Figure~\ref{fig:mismatch_bar} shows, a discrepancy exists for both models, though the gaps are smaller than the Gold model. This implies that even if a ConvQA model is trained on the predictions, the problem of discrepancy should not be ignored. Meanwhile, our proposed models can alleviate such an issue with a selective sampling scheme based on confidence and uncertainty.

\begin{figure}

\centering

\begin{subfigure}[b]{0.235\textwidth}
    \centering
    \includegraphics[width=\textwidth]{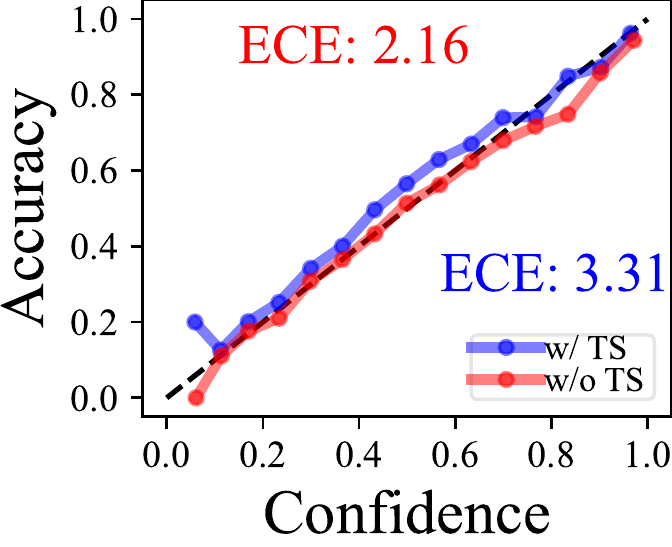}
    %\caption{cap1}
    \label{fig:moti1}
\end{subfigure}
\hfill
\begin{subfigure}[b]{0.235\textwidth}
    \centering
    \includegraphics[width=\textwidth]{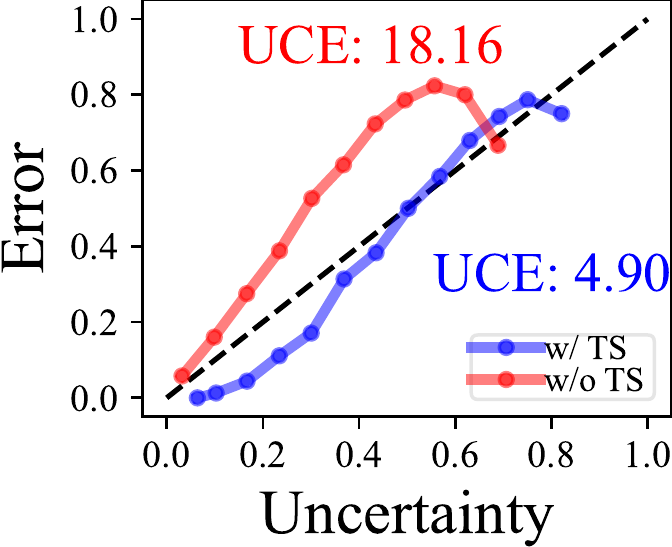}
    %\caption{cap2}
    \label{fig:moti2}
\end{subfigure}
\vspace{-0.18in}
\caption{\small Reliability diagrams with and without temperature scaling (TS), regarding confidence or uncertainty on QuAC.}
\vspace{-0.22in}
\label{fig:ece_uce_diagram}
\end{figure}

\begin{figure}[t!]
\begin{center}
\includegraphics[width=0.49\textwidth]{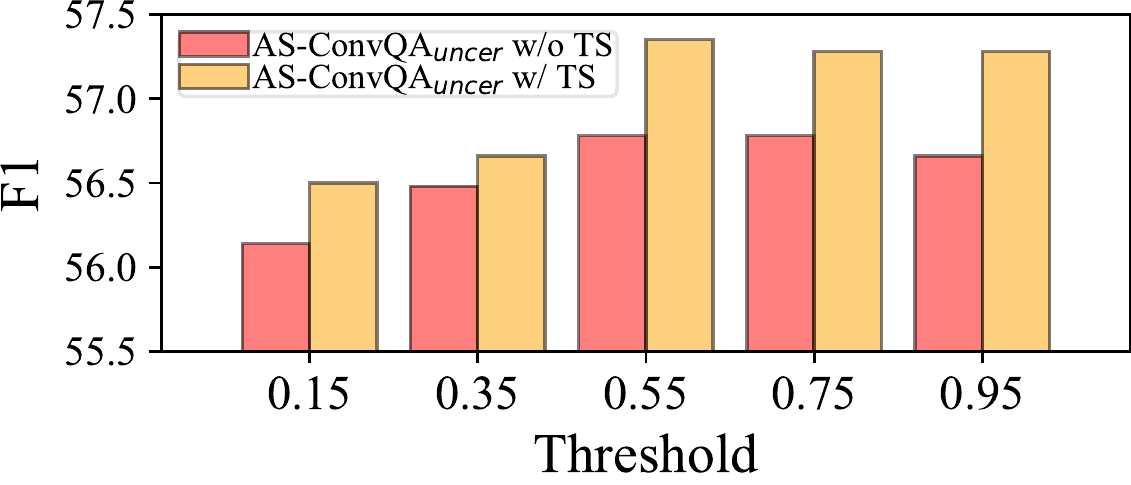}
\end{center}
\vspace{-0.18in}
\caption{
\small Comparison between the calibrated and not calibrated AS-ConvQA$_\texttt{uncer}$ with varying thresholds on QuAC.
}
\label{fig:ablation}
\vspace{-0.00in}
\end{figure}

\paragraph{Effectiveness of Calibration}
We show the effect of calibration on confidence and uncertainty values in Figure~\ref{fig:ece_uce_diagram}. Regarding confidence, the QA model already generates calibrated scores; thus there is no reason to scale the logit vector with temperature scaling (i.e., w/ temperature scaling yields more errors in terms of ECE). However, regarding uncertainty, the estimated uncertainty scores from the model have high errors in terms of UCE, i.e., not calibrated. Thus, after applying the temperate scaling scheme, the uncertainties become calibrated. Note that the calibrated uncertainties further contribute to the performance gain, as Figure~\ref{fig:ablation} shows, since the model can observe a broad range of uncertainty values during training, making the model easily capture and reject uncertain predictions.

\begin{figure}[t!]
\begin{center}
\includegraphics[width=0.49\textwidth]{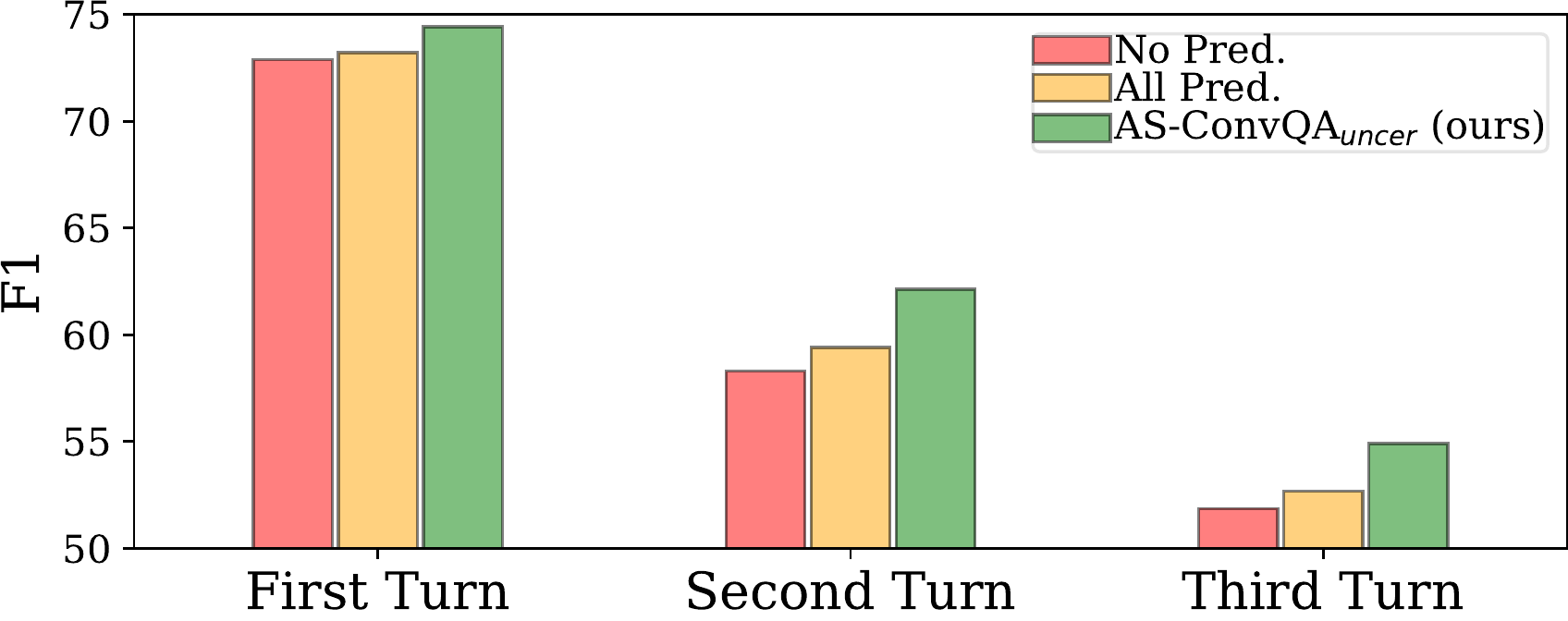}
\end{center}
\vspace{-0.15in}
\caption{
\small F1-scores for the first, second, and third conversational turns on QuAC with baselines and our AS-ConvQA.
}
\label{fig:turn}
\vspace{-0.15in}
\end{figure}

\paragraph{Effectiveness on Conversational Turns}
To see how the proposed AS-ConvQA contributes to the quality of the conversation as it proceeds, we further analyze the performances of former and latter conversational turns. As shown in Figure~\ref{fig:turn}, both the former and latter turns benefit from our AS-ConvQA, and the performance improvements are more significant on the latter turn. This result implies that our AS-ConvQA effectively prevents the accumulation of errors, which originate from the incorrect predictions in the previous turns.

\colorlet{usercolorname}{lightgray!60}
\sethlcolor{usercolorname}

\setstcolor{red}

\begin{table*}[h!]
\small
\centering
\caption{\small Examples in a realistic ConvQA evaluation setting for No Pred., All Pred., and AS-ConvQA$_\texttt{uncer}$ (Ours) models.}
\vspace{-0.075in}
\resizebox{0.99\textwidth}{!}{
\renewcommand{\arraystretch}{1.23}
\begin{tabular}{lcc}
%%%%
\toprule
\midrule

\noalign{\vskip 0.2ex}
\multicolumn{3}{p{\textwidth}}{\normalsize \textbf{Case \# 1:} Our AS-ConvQA removes a previous answer and then predicts a correct answer.} \\
\midrule
\multicolumn{3}{p{\textwidth}}{$\boldsymbol{C_1}$: ... Their collaboration has invited comparisons to the 
collaborations of Steven Spielberg and John Williams. This big break led to Hisaishi's overwhelming success as a composer 
of film scores. In 1986, Laputa: Castle in the Sky, would be the first feature to appear under the Studio Ghibli banner, and \hl{($\boldsymbol{A_6}$) its 
gentle, faintly melancholic tone would become a familiar trademark of much of the studio's later output.} \hl{($\boldsymbol{A_7}$) And later, in the 1990s, 
Porco Rosso and Princess Mononoke were released.}} \\

\multicolumn{3}{p{\textwidth}}{$\boldsymbol{Q_7}$: What other output did the studio release?} \\

\midrule
{} & \textbf{All Pred.} & \textbf{AS-ConvQA$_\texttt{uncer}$ (Ours)} \\
\midrule

\multirow{2}{*}{$\boldsymbol{\mathcal{H}_7}$} 
& \multicolumn{1}{p{0.46\textwidth}}{{ $\boldsymbol{Q_6}$}: What made this so successful?} & \multicolumn{1}{p{0.46\textwidth}}{$\boldsymbol{Q_6}$: What made this so successful?} \\

& \multicolumn{1}{p{0.46\textwidth}}{$\boldsymbol{\bar{A}_6}$: This big break led to Hisaishi's overwhelming success} & \multicolumn{1}{p{0.46\textwidth}}{ \st{$\boldsymbol{\bar{A}_6}$: CANNOTANSWER} \color{red}$\boldsymbol{ (s_{uncer} > \texttt {Threshold})}$} \\

\cdashline{0-2}\noalign{\vskip 0.75ex}

\multirow{2}{*}{$\boldsymbol{\bar{A}_7}$} 
& \multicolumn{1}{p{0.46\textwidth}}{In 1986, Laputa: Castle in the Sky, would be the first feature to appear under the Studio Ghibli banner}  &  \multicolumn{1}{p{0.46\textwidth}}{And later, in the 1990s, Porco Rosso and Princess Mononoke were released.} \\

\midrule
\midrule
%%%%

%\toprule

\noalign{\vskip 0.2ex}
\multicolumn{3}{p{\textwidth}}{\normalsize \textbf{Case \# 2:} Our AS-ConvQA keeps a previous answer and then, based on it, predicts a correct answer.} \\
\midrule
\multicolumn{3}{p{\textwidth}}{$\boldsymbol{C_2}$: ... The Walk, Hanson's second studio album with 3CG Records (Fourth overall), was released in the US, Mexico and Canada on July 24. It was released in Japan on February 21 and in the UK on April 30. On May 6, 2007, the 10th anniversary of Hanson Day, \hl{($\boldsymbol{A_5}$) the band re-recorded their first major label album, Middle Of Nowhere, at The Blank Slate bar in their hometown of Tulsa, Oklahoma.} \hl{($\boldsymbol{A_6}$) The band invited fan club members, causing hundreds to fly to Oklahoma for the acoustic event.} Hanson played concerts in the summer of 2007, supporting release of The Walk.} \\

\multicolumn{3}{p{\textwidth}}{$\boldsymbol{Q_6}$: Was it well received?} \\

\midrule
{} & \textbf{No Pred.} & \textbf{AS-ConvQA$_\texttt{uncer}$ (Ours)} \\
\midrule

\multirow{4}{*}{$\boldsymbol{\mathcal{H}_6}$} 
& \multicolumn{1}{p{0.46\textwidth}}{{$\boldsymbol{Q_5}$}: What did they do on their tenth anniversary?} & \multicolumn{1}{p{0.46\textwidth}}{$\boldsymbol{Q_5}$: What did they do on their tenth anniversary?} \\

& \multicolumn{1}{p{0.46\textwidth}}{\st{$\boldsymbol{\bar{A}_5}$: the band re-recorded their first major label album, Middle Of Nowhere, at The Blank Slate bar in their hometown of Tulsa, Oklahoma.}} & \multicolumn{1}{p{0.46\textwidth}}{ $\boldsymbol{\bar{A}_5}$: the band re-recorded their first major label album, Middle Of Nowhere, at The Blank Slate bar in their hometown of Tulsa, Oklahoma.
\color{blue}$\boldsymbol{ (s_{uncer} < \texttt {Threshold})}$} \\

\cdashline{0-2}\noalign{\vskip 0.75ex}

\multirow{2}{*}{$\boldsymbol{\bar{A}_6}$} 
& \multicolumn{1}{p{0.46\textwidth}}{\multirow{2}{*}{CANNOTANSWER}}  &  \multicolumn{1}{p{0.46\textwidth}}{The band invited fan club members, causing hundreds to fly to Oklahoma for the acoustic event.} \\

\midrule
\midrule

\noalign{\vskip 0.2ex}
\multicolumn{3}{p{\textwidth}}{\normalsize \textbf{Case \# 3:} Our AS-ConvQA predicts an incorrect answer since it filters out a correct previous answer.} \\
\midrule
\multicolumn{3}{p{\textwidth}}{$\boldsymbol{C_3}$: ... Official calendars have also been issued annually from 2004 to 2009, the only exception being 2005. \hl{($\boldsymbol{A_3}$) Girls Aloud co-wrote an autobiography titled Dreams That Glitter - Our Story.} The book, named after a lyric in \"Call the Shots\", was published in October 2008 through the Transworld imprint Bantam Press. Before the release, OK! magazine bought the rights to preview and serialise the book. \hl{($\boldsymbol{A_4}$) In 2007, Girls Aloud signed a PS1.25m one-year deal to endorse hair care brand Sunsilk.} } \\

\multicolumn{3}{p{\textwidth}}{$\boldsymbol{Q_4}$: What else did they do?} \\

\midrule
{} & \textbf{All Pred.} & \textbf{AS-ConvQA$_\texttt{uncer}$ (Ours)} \\
\midrule

\multirow{3}{*}{$\boldsymbol{\mathcal{H}_4}$} 
& \multicolumn{1}{p{0.46\textwidth}}{{ $\boldsymbol{Q_3}$}: What else did they do/create?
} & \multicolumn{1}{p{0.46\textwidth}}{$\boldsymbol{Q_3}$: What else did they do/create?} \\

& \multicolumn{1}{p{0.46\textwidth}}{$\boldsymbol{\bar{A}_3}$: Girls Aloud co-wrote an autobiography titled Dreams That Glitter - Our Story.} & \multicolumn{1}{p{0.46\textwidth}}{ \st{$\boldsymbol{\bar{A}_3}$: Girls Aloud co-wrote an autobiography titled Dreams That Glitter - Our Story.} \color{red}$\boldsymbol{ (s_{uncer} > \texttt {Threshold})}$} \\

\cdashline{0-2}\noalign{\vskip 0.75ex}

\multirow{2}{*}{$\boldsymbol{\bar{A}_4}$} 
& \multicolumn{1}{p{0.46\textwidth}}{In 2007, Girls Aloud signed a PS1.25m one-year deal to endorse hair care brand Sunsilk.}  &  \multicolumn{1}{p{0.46\textwidth}}{Girls Aloud co-wrote an autobiography titled Dreams That Glitter - Our Story.} \\

\midrule
\bottomrule
\end{tabular}
}
\label{tab:case_study}
\vspace{-0.05in}
\end{table*}

\paragraph{Case Study}
We conduct a case study. As the first example in Table~\ref{tab:case_study} shows, even though both All Pred. and AS-ConvQA$_\texttt{uncer}$ models inaccurately predict $\bar {A_6}$, they handle it differently: All Pred. accepts it, while ours reject it for the subsequent question. In particular, All Pred. model misinterprets `this' as `big break' when answering $Q_6$ as well as $Q_7$; however, since `this' actually refers to `Laputa: Castle in the Sky', All Pred. further propagates the misleading prediction to the next question. By contrast, our model decides not to select $\bar {A}_6$ as a conversational history due to its high uncertainty, thus not repeating the previous mistake when answering $Q_7$. The proportion of such examples is about 0.52, where our model predicts the previous answer incorrectly but answers the next question with a high F1-score over 50.
This emphasizes the importance of our selection scheme, especially when there exist ambiguous words prone to mispredictions.

In addition to this case of removing the uncertain previous prediction, we further compare our model against the No Pred. model in the case where the model predicts with the previous answer history having a low uncertainty value. As the second example in Table~\ref{tab:case_study} shows, while both No Pred. and AS-ConvQA$_\texttt{uncer}$ correctly predict $\bar A_5$, No Pred. does not use $\bar A_5$ as the answer history when answering the next question, $Q_6$. However, as $\bar A_5$ contains important information of `it' in $Q_6$, No Pred. model gives an inaccurate answer to $Q_6$, since the model is confused about what `it' refers to. On the other hand, our model selects $\bar A_5$ as the answer history due to its low uncertainty value, thereby correctly predicting $\bar A_6$ with the previous prediction $\bar A_5$. In a third example of Table~\ref{tab:case_study}, we show the potential failure of our model, which is discussed in the Limitations section after Section~\ref{sec:limitation}.

\section{Conclusion}
\vspace{-0.050in}

In this work, in order to tackle the challenge of inaccurately predicted answers in the conversation history, we proposed a novel answer selection scheme based on their confidence and uncertainty values. We further calibrated the output values of the model to match the model's predicted confidence and uncertainty to its correct likelihood and error, which makes our answer selection scheme more reliable. The experimental results and analyses demonstrate that AS-ConvQA significantly improves the ConvQA model performance in a realistic evaluation setting without making any architectural changes.

\section*{Limitations}
\label{sec:limitation}

\vspace{-0.050in}

While we show the clear advantages of using our AS-ConvQA in realistic ConvQA tasks with both quantitative and qualitative perspectives, there could be possible failures: estimated confidence and uncertainty of a model's prediction do not match its actual correctness.
For instance, the third example in Table~\ref{tab:case_study} shows that AS-ConvQA$_\texttt{uncer}$ gives an incorrect answer to $Q_4$, since it removes the correctly predicted previous answer (i.e., $\bar A_3$) due to its incorrectly estimated uncertainty. Specifically, both $Q_3$ and $Q_4$ ask the additional information: `What else did they do?'. However, the erroneous deletion of $\bar A_3$ makes our model bound to the previous question, repeatedly giving the same answer as $\bar A_3$. This implies that AS-ConvQA$_\texttt{uncer}$ sometimes assigns high uncertainty to the correct prediction and filters it, which may mislead the model, especially for the one that requires careful attention to the context with the previous answer. Therefore, as future work, one may improve mechanisms to measure incorrectness of predictions.

\vspace{-0.030in}
\section*{Ethics Statement}

\vspace{-0.050in}

As the need for fully autonomous conversational agents has been rapidly emerging, it is crucial to consider whether ConvQA models can correctly answer a sequence of questions in a realistic setting, in which gold answers for previous questions are unavailable. We note that, in such a challenging setting, our work contributes to the improved performance by selectively using predicted answers with model confidence and uncertainty instead of using predefined gold answers. However, as ConvQA models predict answers based on the given paragraph, we should further consider a scenario where the paragraph itself is not trustworthy, sometimes having offensive contents. Subsequently, this may lead the entire conversation vulnerable to generating unexpected and undesired texts. While this is not the concern raised from our proposed AS-ConvQA models themselves, we still have to make an effort to prevent such an undesirable behavior.

\vspace{-0.030in}
\section*{Acknowledgements}
\vspace{-0.050in}

This work was supported by Institute for Information and communications Technology Promotion (IITP) grant funded by the Korea government (MSIT) (No. 2018-0-00582, Prediction and augmentation of the credibility distribution via linguistic analysis and automated evidence document collection).

\bibliography{anthology,custom}
\bibliographystyle{acl_natbib}

\clearpage

\appendix

\section{Experimental Implementation Details}
\label{appendix:implementation_details}

We implement all models using PyTorch~\cite{pytorch} and Transformers library~\cite{wolf-etal-2020-transformers}. For language models, we use BERT-base and RoBERTa-base models with 110M and 125M parameters, respectively. For training, we set the training epoch as 2 with the batch size of 12, where the first epoch is used for Step 1 while the second epoch is used for Step 2. Furthermore, we optimize all models with the Adam optimizer~\cite{DBLP:journals/corr/KingmaB14} with a learning rate of 3e-5. For computiting resources, we use a single GeForce RTX 3090 GPU with 24GB memory, on which each training epoch requires approximately 4 hours. 

For hyperparameters, we search the temperature value $\tau$ for temperature scaling with a validation set, in the range of $(0,2]$. Also, we set the filtering threshold for Step 2, in the range of $[\texttt{median}-0.25, \texttt{median}+0.25]$, where $\texttt{median}$ is the median value of confidence or uncertainty for all samples. For the number of dropout masks (i.e., $N$ for the uncertainty estimation in Section~\ref{subsec:selection}) for measuring uncertainty, we set it as $10$.

We use two benchmark ConvQA datasets, which are QuAC\footnote{https://quac.ai/}~\cite{convqa1} and CoQA\footnote{https://stanfordnlp.github.io/coqa/}~\cite{convqa2}.
Note that, while our main focus is on predicting the extractive answers within a given context, CoQA is designed for answering question in a free-form text, which might not appear in a given context. Therefore, following the experimental setting from~\citet{convqa2}, we convert the CoQA dataset to our extractive ConvQA setting. In particular, we assume the gold answer as the provided rationale, and then make prediction on it, except for simple yes or no questions. For the yes or no questions, we additionally augment yes and no tokens at the end of the paragraph.

\section{Additional Experimental Results}
\label{appendix:results}

\begin{figure}[t!]
\begin{center}
\includegraphics[width=0.48\textwidth]{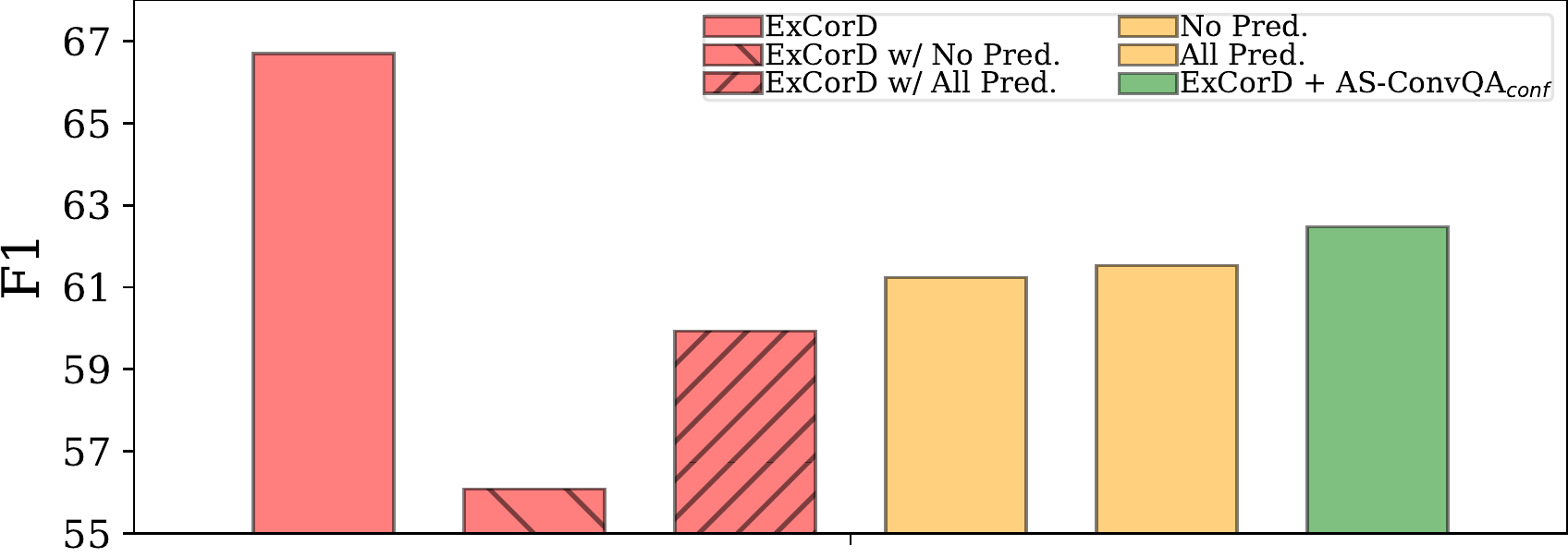}
\end{center}
\vspace{-0.13in}
\caption{
\small F1-scores on the mismatching evaluation settings for the recent ExCorD model~\cite{excord} on QuAC.
}
\label{fig:excord}
\vspace{-0.18in}
\end{figure}

\paragraph{Realistic Evaluation of ExCorD}
Even though we validate a negative impact of exposure bias in Figure~\ref{fig:mismatch_bar}, we further explore the performance of the unrealistic state-of-the-art model, ExCorD~\cite{excord}, that uses gold answer histories, in Figure~\ref{fig:excord} with the realistic ConvQA setting that uses predicted answers. We observe that, similar to the mismatching evaluation experiments reported in Figure~\ref{fig:mismatch_bar}, the F1-scores of ExCorD drastically drop when evaluated with No Pred. and All Pred. settings, which aligns with our motivation. On the other hand, the performance is much improved by further adapting our AS-ConvQA on ExCorD. This result indicates the importance of filtering unnecessary predictions together with the applicability of our AS-ConvQA model in a realistic setting.

\begin{figure}[t]
\begin{center}
\includegraphics[width=0.48\textwidth]{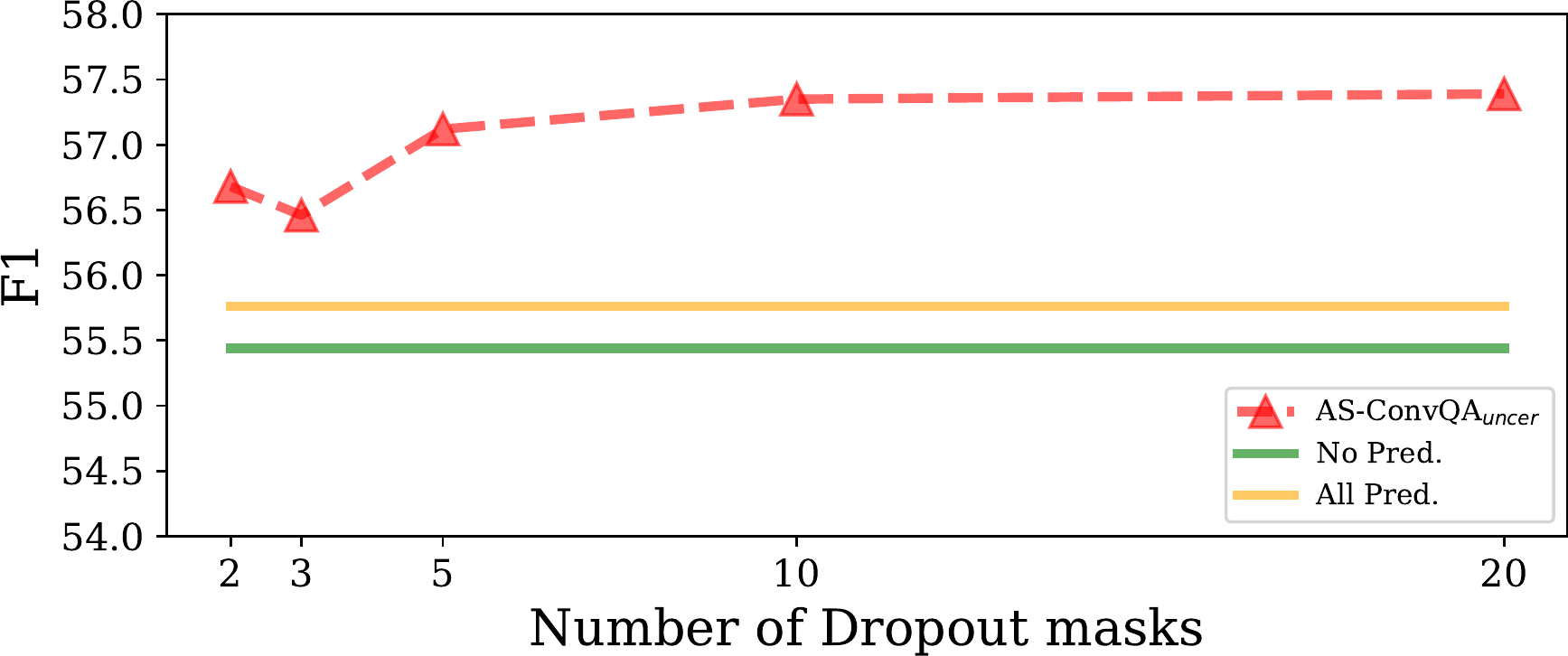}
\end{center}
\vspace{-0.13in}
\caption{
\small F1 scores with varying dropout numbers on QuAC.
}
\label{fig:dropout_mask_line}
\vspace{-0.1in}
\end{figure}

\paragraph{Varying the Number of Dropout Masks}
In order to understand how the number of dropout masks (i.e., $N$ used for uncertainty estimation in Section~\ref{subsec:selection}) affects the performance, we vary the number of masks for AS-ConvQA$_\texttt{uncer}$. As Figure~\ref{fig:dropout_mask_line} shows, the performance is stabilized after a certain number of masks (i.e., 5). This indicates the importance of setting an appropriate sampling number, since approximating the uncertainty with a small number of masks is likely to be inaccurate.

\end{document}